\documentclass[runningheads]{llncs}

\usepackage[T1]{fontenc}
\def\doi#1{\href{https://doi.org/\detokenize{#1}}{\url{https://doi.org/\detokenize{#1}}}}
\usepackage{color}
\usepackage{hyperref}

\hypersetup{hypertex=true,
            colorlinks=true,
            linkcolor=blue,
            anchorcolor=blue,
            citecolor=blue}

\usepackage{cite}
\usepackage{graphicx}
\usepackage{multirow}
\usepackage[misc]{ifsym}
\usepackage{fontawesome}

\usepackage{listings}
\begin{document}
\title{Out-of-Distribution Detection for Long-tailed and Fine-grained Skin Lesion Images}
\titlerunning{Out-of-Distribution Detection for Skin Lesion Images}
\author{Deval, Mehta\inst{1,2}\textsuperscript{(\faEnvelopeO)}
\and Yaniv Gal\inst{3} \and Adrian Bowling\inst{3} \and Paul Bonnington\inst{2} \and Zongyuan Ge\inst{1,2,4}}
%index{Mehta, Deval}
%index{Gal, Yaniv}
%index{Bowling, Adrian}
%index{Bonnington, Paul}
%index{Ge, Zongyuan}
%
\authorrunning{D. Mehta et al.}
\institute{Monash Medical AI, Monash University, Melbourne, Australia \url{https://www.monash.edu/mmai-group} \\ \email{deval.mehta@monash.edu} \and 
eResearch Centre, Monash University, Melbourne, Australia \\ \and
Kahu AI, Auckland, NewZealnd \\ \and
Monash Airdoc Research, Monash eResearch Centre, Melbourne, Australia}
\maketitle              % typeset the header of the contribution
\begin{abstract}
Recent years have witnessed a rapid development of automated methods for skin lesion diagnosis and classification. Due to an increasing deployment of such systems in clinics, it has become important to develop a more robust system towards various Out-of-Distribution (OOD) samples (unknown skin lesions and conditions). However, the current deep learning models trained for skin lesion classification tend to classify these OOD samples incorrectly into one of their learned skin lesion categories. To address this issue, we propose a simple yet strategic approach that improves the OOD detection performance while maintaining the multi-class classification accuracy for the known categories of skin lesion. To specify, this approach is built upon a realistic scenario of a long-tailed and fine-grained OOD detection task for skin lesion images. Through this approach, 1) First, we target the mixup amongst middle and tail classes to address the long-tail problem. 2) Later, we combine the above mixup strategy with prototype learning to address the fine-grained nature of the dataset. The unique contribution of this paper is two-fold, justified by extensive experiments. First, we present a realistic problem setting of OOD task for skin lesion. Second, we propose an approach to target the long-tailed and fine-grained aspects of the problem setting simultaneously to increase the OOD performance.

\keywords{skin lesion  \and out-of-distribution \and openset \and mixup \and prototype.}
\end{abstract}
\section{Introduction}
Early detection and diagnosis of skin cancer remains a global challenge. Following the advent and success of deep learning for various computer vision tasks, medical research community has seen many deep learning models being developed for detection and classification of skin cancer~\cite{esteva2017dermatologist}. While many models~\cite{gessert2020skin,mahbod2020transfer} have been developed to achieve a high performance, their evaluation and validation is limited to small datasets~\cite{codella2018skin,tschandl2018ham10000} which have limited shifts and variations in input data. Particularly, if such models are used in real practice, they are bound to encounter many Out-of-Distribution (OOD) samples which may represent unknown skin conditions, hardware device variations, and different clinical settings. Since these OOD samples are not in their training set, these models may assign a high confidence score and classify them as one of the known categories of their training set. Thus, it becomes vital to build the model's capability of rejecting the unknown OOD samples and make them robust for practical deployment.

\begin{figure}[t!]
\centering
\includegraphics[keepaspectratio,width=0.87\textwidth]{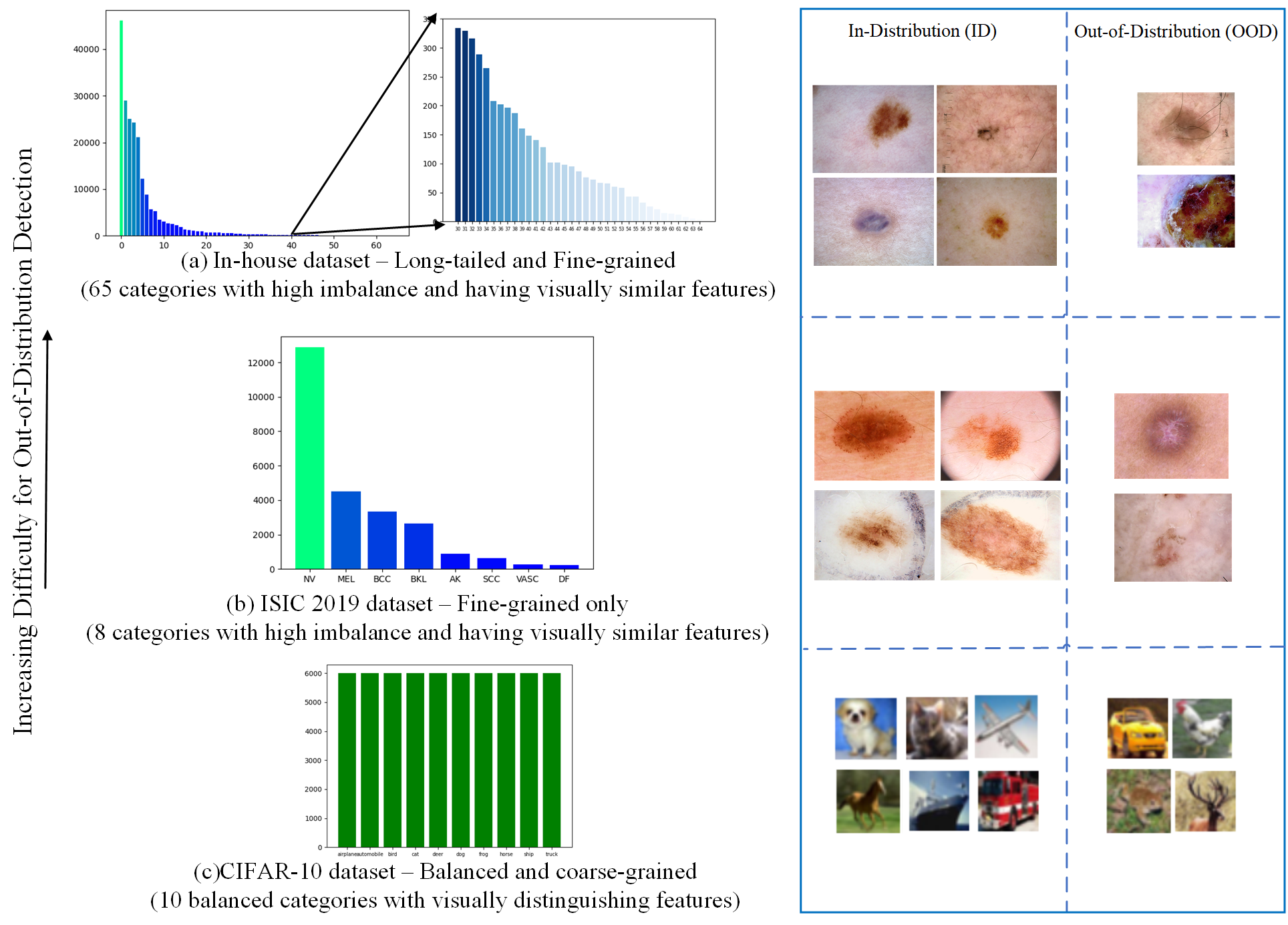}
\caption{Difficulty of Out-of-Distribution (OOD) task for different datasets (a) Our in-house long-tailed and fine-grained dataset, (b) ISIC 2019 fine-grained dataset, (c) CIFAR-10 dataset - Balanced and coarse-grained. (x: Classes; y: Number of samples).} \label{fig1}
\end{figure}

The OOD detection task and openset recognition (OSR) problem was formalized by~\cite{6365193}. Since then, it has received much needed attention from the research community in general computer vision~\cite{geng2020recent}. There have been many methods proposed with diverse strategies ranging from scaling the output softmax scores~\cite{bendale2016towards} (OpenMax), energy-based separation~\cite{liu2020energy} and perturbed input pre-processing~\cite{hendrycks2018deep} (MSP). Techniques such as ODIN~\cite{liang2017enhancing} have become standard baselines for separating the OOD samples based on the scaled softmax score. Another category of work concentrated on supplying synthetic samples generated from GANs for making the classification model aware of the OOD boundaries~\cite{lee2017training,ge2017generative,chen2021adversarial}. Other directions include using reconstruction based errors~\cite{yoshihashi2019classification,sun2020conditional} for separating closed set and OOD samples, and using self-supervised learning~\cite{hendrycks2019using,tack2020csi} for better learning of closed set feature space. There have also been some recent works~\cite{budhwant2020open,kim2021out,combalia2020uncertainty,pacheco2020out,roy2022does} based on the above ideas for developing OOD detection techniques for skin lesions utilizing the ISIC dataset.

Although the research community has made good contributions recently for the OOD detection task, we want to highlight that the commonly used problem settings and datasets for performance evaluation do not resemble a scenario for clinical deployment purposes. There are two shortcomings in this aspect - 1) The closed set dataset used for training the model is extremely well balanced and coarse-grained in nature for e.g. CIFAR-10~\cite{krizhevsky2009learning}, which have visually distinguishable features between their categories (e.g. dog, car, plane etc. as shown in Fig~\ref{fig1}(c)). This makes the detection of OOD samples relatively easy as they are some of the reserved categories from the dataset. 2) Even if a fine-grained dataset is selected, its distribution is usually not long-tailed in nature. An example for that is the ISIC~\cite{codella2018skin} dataset which only contains a handful number of categories. These settings thus do not resemble a real-world application scenario where there are significantly higher number of fine-grained categories with a long-tailed distribution like our in-house dataset shown in Fig~\ref{fig1}(a).

In this work, we conduct our OOD detection study on an in-house collected dataset (Molemap) from a clinical environment following tele-dermatology labelling standards. The dataset includes 208,287 tele-dermatology verified dermoscopic images categorized into 65 different skin conditions making it a long-tailed and fine-grained skin lesion dataset as shown in Fig.\ref{fig1}(a). The long-tailed and fine-grained challenges of our dataset inspires us to develop a method for targeting both these aspects to achieve a better OOD performance compared to the existing techniques. Specifically, we develop an inter-subset mixup strategy for targeting the middle and tail classes and combine it with prototype learning to tackle the fine-grained aspect. We conduct extensive experiments on both our in-house dataset and ISIC2019 dataset for benchmarking and validation.

\section {Proposed Method}
Learning better decision boundaries between closed set categories helps the classifier for detecting OOD samples more accurately~\cite{vaze2021open}. This learning is dependent on the nature of a dataset. For a long-tailed dataset, head classes dominate the learning process as they contain a large number of samples. Although random oversampling/undersampling and techniques like SMOTE~\cite{chawla2002smote} can be used to tackle this problem, repeating/removing samples of classes does not help the classifier learn any better decision boundaries. Instead, our proposed approach employs a combination of data augmentation using mixup and better feature space learning using prototype loss specifically targeted to middle and tail classes. This enables us to improve the classification performance for those middle and tail categories which also increases the OOD detection performance. Our simple yet strategic method combines two established techniques to target the long-tailed and fine-grained aspects of our challenging dataset.

\begin{figure}
\centering
\includegraphics[keepaspectratio,width=0.8\textwidth]{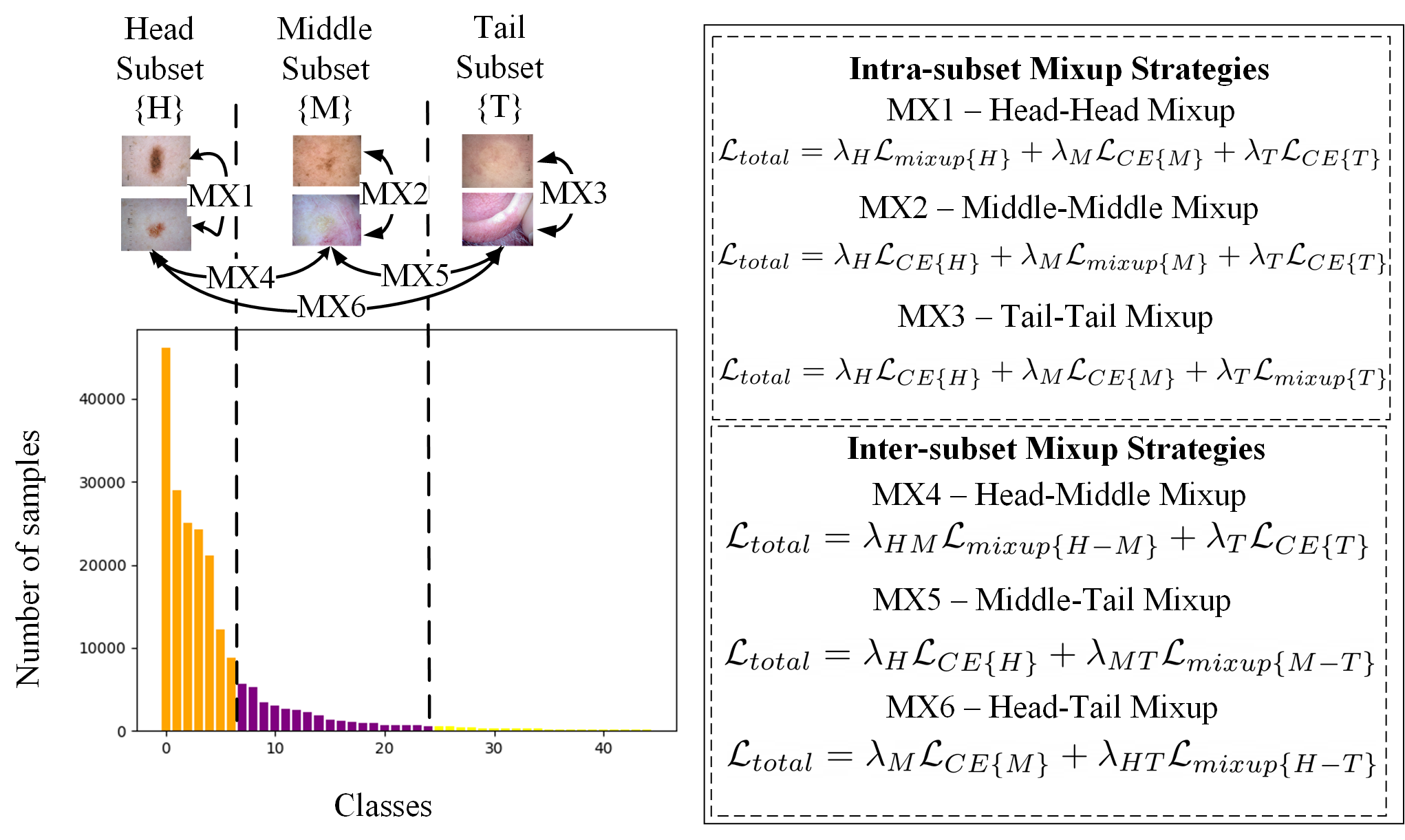}
\caption{Proposed Mixup Strategies - Intra-subset and Inter-subset} \label{fig2}
\end{figure}

\subsection{Intra-subset and Inter-subset Mixup Strategies}
The approach of Mixup~\cite{zhang2017mixup} has become quite popular in data augmentation space where it combines a pair of samples and their labels. This simple technique has been shown to increase robustness towards adversarial samples, a better estimate of uncertainty~\cite{thulasidasan2019mixup}, and better generalization of the trained model~\cite{zhang2020does}. The Mixup sample helps the model training to differentiate between the two categories in its sample even more and thus creates better decision boundaries between the two categories. We believe that Mixup strategy for skin lesion images will provide a better latent space representation between the different categories. However, simply adopting the Mixup strategy for a long-tailed dataset will still be heavily influenced by the head classes thus limiting its advantages only to a small part of the dataset.

To mitigate this problem, we devise different strategies for adopting mixup targeted to a certain part of the dataset only. We partition the total categories set ($C$) into three subsets based on the number of samples - Head ($H \subset C$), Middle ($M \subset C$), and Tail ($T \subset C$) as shown in Fig~\ref{fig2}. With this partition, we devise three intra-subset and three inter-subset mixup strategies. As the name suggests, in an intra-subset strategy, both the independent category samples belong to the same subset (for e.g. in MX1 strategy, both the independent samples will come from the categories in the $\{H\}$ subset). Whereas, in an inter-subset strategy, one sample comes from one of the categories of that subset and the other sample likewise (for e.g. in MX4 strategy, one sample comes from subset $\{H\}$ and the other sample comes from subset $\{M\}$). Thus, we can have six such strategies MX1 to MX6 as depicted in Fig~\ref{fig2}.
The network training strategy is changed accordingly based on the mixup strategy adopted. It is updated with the combined loss of cross-entropy for the conventional subset learning and the mixup loss (given by Eq~\ref{eq1}) for the specific subset selected for mixup. In our experiments, we find that targeting the mixup only amongst the middle and tail classes is effective for achieving better OOD detection performance.

\begin{equation}
\mathcal{L}_{mixup}=\lambda  \mathcal{L}_{CE}(f(x_{i}),y_{i}) +(1-\lambda) \mathcal{L}_{CE}(f(x_{j}),y_{j})
\label{eq1}
\end{equation}

\noindent where $\{x_i$,$x_j\}$ are the input samples of class $\{C_i,C_j\}$ having labels $\{y_i,y_j\}$ respectively, $\lambda$ is the weightage parameter for each sample between \{0,1\}.

\begin{figure}[t!]
\centering
\includegraphics[width=0.8\textwidth]{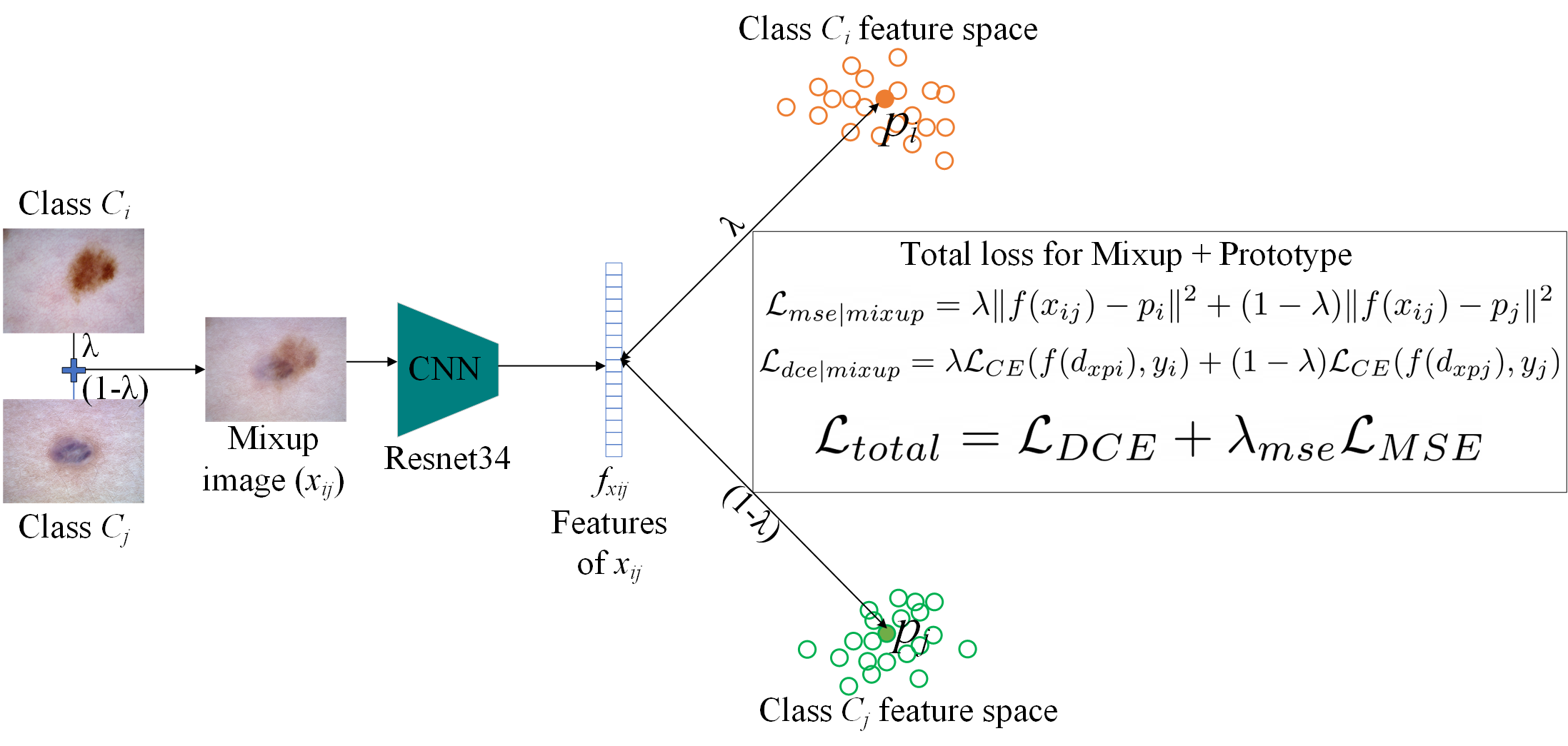}
\caption{Proposed combination of Mixup and Prototype Learning} \label{fig3}
\end{figure}

\subsection{Integration of Mixup with Prototype Learning}
\cite{yang2018robust} introduced prototype learning for increasing the robustness of classifiers by proposing the prototype loss for reducing the intra-class distance variation and increasing the inter-class distance, which has proved extremely effective for learning fine-grained features. A standard prototype loss consists of the mean squared error loss and distance based cross entropy loss between the latent features from an encoder and the corresponding prototypes of all the categories. In our proposed approach, we further \textit{enhance} the capability of prototype learning by combining it with mixup. The framework for this integration is shown in Fig~\ref{fig3}. Firstly, a mixup sample is created from two independent samples belonging to different class categories $\{C_i,C_j\}$ with weights $\lambda$ and $(1-\lambda)$ respectively. This is then fed to a deep neural network (CNN) to generate its feature embeddings $f_{xij}$. As this input is a combination of a mixup of two different classes with a specific weightage, we propose to calculate the distance of the features $f_{xij}$ from the class specific prototypes $\{p_i,p_j\}$ based on the weightage parameter $\lambda$ and also do the same for the distance-based cross-entropy loss. With this, we design a custom loss function for the mixup sample combined with prototype learning given by Eqs~\ref{eq2} and~\ref{eq3}, which is adopted for the mixup samples only. For the normal samples, we adopt the standard prototype loss. Finally, we combine our best performing mixup strategy with prototype learning to address both the long-tailed and fine-grained aspects of our dataset.

\begin{equation}
\mathcal{L}_{mse|mixup}=\lambda\|f(x_{ij}) - p_{i}\| ^{2}+(1-\lambda)\|f(x_{ij}) - p_{j}\|^{2}
\label{eq2}
\end{equation}

\begin{equation}
\mathcal{L}_{dce|mixup} = \lambda\mathcal{L}_{CE}(d_{xpi},y_i) + (1-\lambda)\mathcal{L}_{CE}(d_{xpj},y_j)
\label{eq3}
\end{equation}

\noindent where $\{d_{xpi},d_{xpj}\}$ is the square of the distance between the feature $f_{xij}$ from the class specific prototypes $\{p_i,p_j\}$. The total loss is depicted in Fig~\ref{fig3}.

\section{Experimental Results}
\subsection{Dataset Settings and Evaluation Metrics}
\subsubsection{In-house and ISIC 2019 dataset}
For evaluation purposes, we do extensive experiments on our in-house dataset and ISIC 2019 dataset. For our in-house dataset, we partition the 65 categories into 6 Head (more than 10,000 samples), 17 Middle (500 to 10,000 samples), 22 Tail (less than 500), and the rest 20 are reserved as OOD categories. As we wish to replicate a realistic clinical setting, we select this split where most samples are concentrated in the head categories, with a moderate number of samples in the middle and the least number of samples in the tail categories. Thus, we have 45 ID categories and 20 OOD categories. For the 45 ID categories, we split the data into 85\%-15\% train-test split. The train set is further divided into 80\%-20\% for training and validation. We utilize our in-house dataset for selecting all the hyperparameters via grid search (supplementary material) and use those for ISIC 2019 dataset as well.
ISIC 2019 dataset has a class distribution as - MEL-4522, NV-12875, BCC-3323, AK-867, BKL-2624, DF-239, VASC-253, SCC-628. Thus, we select DF and VASC as the OOD categories. The 6 ID categories are separated as Head (NV \& MEL), Middle (BCC \& BKL), and Tail (AK \& SCC) as we wish to have at least two categories in a subset. We adopt the five-fold cross-validation for the ID part of the ISIC 2019 dataset.
\subsubsection{CIFAR-10 and unusual OOD samples}
For a comprehensive study on different types of OOD samples, we add 1000 more OOD samples distributed equally amongst the 10 categories from the CIFAR-10 test set. We further enhance our OOD testing set with some commonly encountered \textbf{unusual} images in a clinic such as blurred images of skin lesions and ones that are completely covered by hair, ear, etc. A good model should give a low confidence score for these samples.
\subsubsection{Training Implementation and Evaluation Metrics}
We use Resnet34~\cite{he2016deep} as our backbone architecture and train all the strategies using Adam optimizer with a batch size of 32, an initial learning rate of 1e-4 with exponential decay for 45 epochs. We resize the input image to a size of 224x224 and adopt the standard data augmentation of random crop and horizontal flip. Our proposed approach is trained only on the ID categories and we \textbf{do not} utilize OOD samples for fine-tuning our hyperparameters. We use precision (\textbf{pre}), recall (\textbf{rec}), and f1-score (\textbf{f1}) as the evaluation parameters for the closed set performance and Area Under Receiver Operator Characteristic (\textbf{AUROC}) as the OOD detection metric, which are the standard metrics for measuring the performance of a model for OOD detection task~\cite{geng2020recent}. Our implementation code is available here\footnote{\url{https://github.com/DevD1092/ood-skin-lesion}}. 
\subsection{Ablation Study of Mixup Strategies}

Table~\ref{tab1} shows the experimental results of the six proposed mixup strategies for our in-house dataset. We show the closed set accuracy for the head, middle and tail subsets separately. We also depict the corresponding OOD performance on the 20 OOD classes of our in-house dataset. Firstly, it can be noted that compared to the standard baseline, simply adopting the Mixup strategy increases the overall closed set performance as well as OOD performance by ~3.3\% and ~0.7\%, however, we want to highlight that the closed set accuracy of tail classes reduces. Secondly, compared to the standard mixup, intra-subset mixup strategies - MX1, MX2, MX3 help to increase the corresponding subset closed set accuracies. From the intra-subset experiments, it should be specifically noted that MX2 and MX3 help to increase the OOD performance suggesting that targeting the middle and tail classes is important for a better OOD detection. For the inter-subset mixup strategies - MX4, MX5, M6, we note that MX5 significantly increases the OOD performance by ~3\%, which suggests inter-subset mixup is more effective compared to the intra-subset strategies for middle and tail categories. It should also be noted that although the OOD performance increases for MX5, the overall accuracy only increases slightly when compared to the baseline. This is due to the reduced influence of the head subset of the dataset. Thus, we use the setting of MX5 for integration with the prototype learning as our final framework strategy.

\begin{table}[t!]
\centering
\caption{Performance evaluation of proposed mixup strategies on our in-house dataset}\label{tab1}
\resizebox{0.75\textwidth}{!}{\begin{tabular}{c|c|c|c|c|c}
\hline
\multirow{2}{*}{Mixup Strategy}&\multicolumn{4}{c|}{Closed set (ID) (Acc\%)} &\multirow{2}{*}{OOD (AUROC\%)}\\
\cline{2-5}
 & Head & Middle & Tail & Total & \\
\hline
Baseline & 66.67 & 38.26 & 36.98 & 60.56 & 65.67 \\
Standard Mixup & 67.23 & 45.18 & 33.89 & \textbf{63.90} & 66.35 \\
\hline
H-H Intrasubset (MX1) & \textbf{70.11} & 34.74 & 25.14 & 60.84 & 64.18 \\
M-M Intrasubset (MX2) & 63.12 & \textbf{55.36} & 31.54 & 61.80 & 66.47 \\
T-T Intrasubset (MX3) & 64.29 & 47.96 & \textbf{39.49} & 61.06 & 66.25 \\
\hline
H-M Intersubset (MX4) & 66.92 & 44.97 & 22.21 & 62.31 & 64.33 \\
\textbf{M-T Intersubset (MX5)} & 63.67 & 55.14 & 38.76 & 60.97 & \textbf{68.78} \\
H-T Intersubset (MX6) & 66.95 & 36.32 & 36.67 & 59.24 & 64.45  \\
\hline
\end{tabular}}
\end{table}

\subsection{Benchmarking with other methods}

\begin{table}
\caption{Benchmarking of OOD techniques on both In-house and ISIC2019 dataset. ID metrics -\{Precision (pre), Recall (rec), and f1-score(f1)\}; OOD metrics - \{AUROC(\%)\} (best viewed in zoom).}  \label{tab2}
\resizebox{\textwidth}{!}{\begin{tabular}{c|c|c|c|c|c|c|c|c|c|c|c}
\hline
\multirow{2}{*}{Method}&\multicolumn{6}{c|}{In-house dataset} &\multicolumn{5}{|c}{ISIC2019}\\
\cline{2-12}
 & ID(pre) & ID(rec) & ID(f1) & OOD(20cl) & OOD(unk) & OOD(Cifar) & ID(pre) & ID(rec) & ID(f1) & OOD(2cl) & OOD(Cifar) \\
\hline
Baseline & 0.58 & 0.59 & 0.585 & 65.67 & 52.90 & 73.24 & $0.86 \pm 0.03$ & $0.86 \pm 0.02$ & $0.86 \pm 0.02$ & $68.15 \pm 0.9$ & $76.43 \pm 0.4$ \\
Baseline+LS+RandAug+LRS~\cite{vaze2021open} & 0.62 & 0.63 & 0.625 & 66.19 & 63.13 & 96.34 & $0.87 \pm 0.015$ & $0.86 \pm 0.017$ & $0.865 \pm 0.015$ & $69.41 \pm 0.5$ & $94.87 \pm 0.6$ \\
ODIN~\cite{liang2017enhancing} & 0.61 & 0.59 & 0.60 & 64.92 & 62.79 & 96.48 & $0.83 \pm 0.03$ & $0.81 \pm 0.02$ & $0.82 \pm 0.025$  & $66.21 \pm 1.3$ & $95.60 \pm 1.2$ \\
OLTR~\cite{liu2019large} & 0.63 & 0.62 & 0.625 & 67.42 & 70.72 & 98.00 & $0.85 \pm 0.01$ & $0.86 \pm 0.02$ & $0.855 \pm 0.015$ & $71.66 \pm 0.6$ & $98.45 \pm 0.5$  \\
MC-Dropout~\cite{gal2016dropout} & 0.59 & 0.58 & 0.585 & 66.07 & 68.83 & 97.57 & $0.84 \pm 0.023$ & $0.84 \pm 0.02$ & $0.84 \pm 0.02$ & $72.18 \pm 0.3$ & $96.41 \pm 0.3$ \\
ARPL~\cite{chen2021adversarial} & \textbf{0.64} & \textbf{0.63} & \textbf{0.635} & 68.55 & 80.61 & 99.42 & $0.85 \pm 0.01$ & $0.86 \pm 0.016$ & $0.855 \pm 0.012$ & $74.16 \pm 0.7$ & $97.20 \pm 0.4$ \\
\hline
Mixup~\cite{zhang2017mixup} & 0.63 & 0.62 & 0.625 & 66.35 & 66.70 & 97.10 & $\textbf{0.87} \pm 0.02$ & $\textbf{0.88} \pm 0.01$ & $\textbf{0.875} \pm 0.013$ & $71.72 \pm 0.7$ & $96.65 \pm 0.6$ \\
Prototype~\cite{yang2018robust} & 0.63 & 0.62 & 0.625 & 68.82 & 74.54 & 98.04 & $0.85 \pm 0.02$ & $0.86 \pm 0.02$ & $0.855 \pm 0.02$ & $72.84 \pm 0.6$ & $97.02 \pm 0.5$ \\
\textbf{M-T Mixup (Ours)} & 0.61 & 0.60 & 0.605 & 68.78 & 70.81 & 99.29 & $0.85 \pm 0.03$ & $0.85 \pm 0.02$ & $0.85 \pm 0.022$ & $73.86 \pm 0.6$ & $97.10 \pm 0.6$ \\
\textbf{M-T Mixup + Prototype (Ours)} & 0.62 & 0.61 & 0.615 & \textbf{71.10} & \textbf{82.71} & \textbf{99.59} & $0.85 \pm 0.01$ & $0.86 \pm 0.02$ & $0.855 \pm 0.015$ & $\textbf{76.37} \pm 0.5$ & $\textbf{98.46} \pm 0.4$ \\
\hline
\end{tabular}}
\end{table}

\noindent In Table~\ref{tab2} we show the performance of all the existing OOD techniques and compare it to our proposed strategies both on our in-house dataset and ISIC2019 dataset. For our in-house dataset, we also have some unusual OOD samples depicted by OOD(unk). We would firstly like to highlight two important observations from this evaluation - 1) It can be seen that all the OOD techniques perform significantly better in detecting CIFAR-10 OOD samples with OOD performance >95\%. This shows that the existing OOD techniques are capable of detecting relatively easy OOD samples coming from a completely different domain. 2) The performance of all the techniques drops drastically when the OOD samples are from the same domain. Specifically, this is more evident for a long-tailed nature dataset such as our in-house dataset where the performance degradation is more severe compared to that of the ISIC2019. Moreover, for our in-house dataset, the OOD performance is slightly better for the unusual OOD(unk) images compared to those reserved from the extreme tail part OOD(20cl). It can be further noted that our proposed approach of M-T mixup (MX5) strategy combined with prototype learning performs the best for OOD detection while maintaining the overall ID performance compared to the baseline on both datasets.

\subsection{Confidence Scores Visualization}

\begin{figure}[t!]
\centering
\includegraphics[keepaspectratio,width=0.95\textwidth]{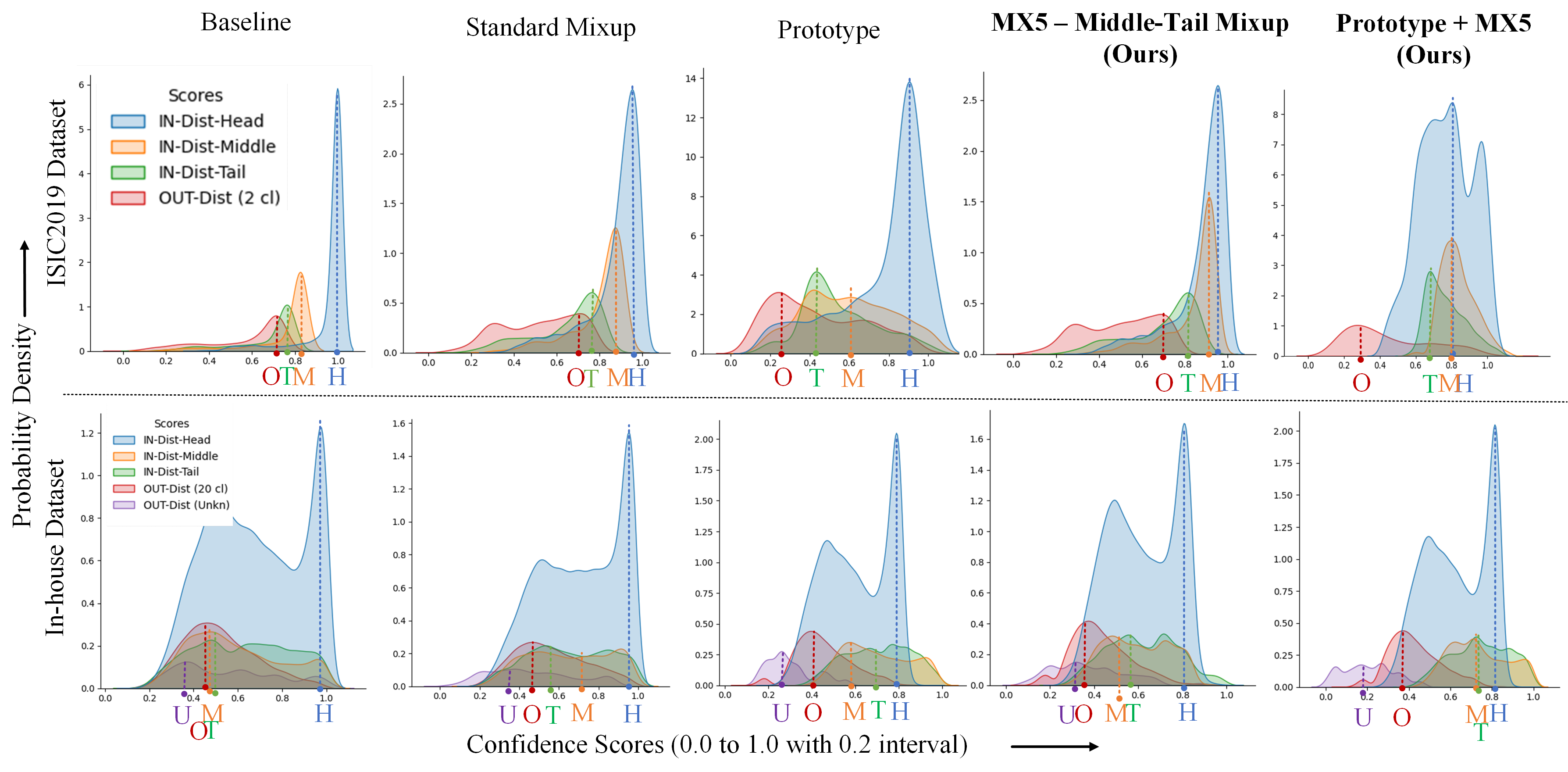}
\caption{Confidence Scores visualization for different methods on our In-house dataset and ISIC dataset. \{H,M,T\} refer to Head, Middle, and Tail subsets. \{O\} refers to OOD(2cl) and OOD(20cl) for ISIC and In-house dataset. \{U\} refers to the OOD(unk) for In-house dataset. (best viewed in zoom).} \label{fig4}
\end{figure}

In Fig~\ref{fig4}, we analyse the performance results in more detail by showing the probability density of the confidence scores for different subsets. The larger the separation between the distribution of OOD$\{O,U\}$ from ID$\{H,M,T\}$, the better the technique. It can be seen that our proposed approach achieves the largest separation for both the datasets. Specifically, it is to be noted that this is achieved by making the \{M,T\} subsets more confident which justifies our targeted strategy.

\section{Conclusion}
In this work, we present a detailed study and a strategic method for out-of-distribution detection for skin lesion images. Several vital conclusions can be made from our extensive experimentation. Firstly, we highlight that the current OOD techniques are still far away from clinical deployment where they will encounter many similar domain OOD images. To fill this gap, we propose a simple combination of middle-tail subset targeted mixup with prototype learning which is necessary for increasing the OOD performance for a long-tailed and fine-grained dataset. We believe that the experimental settings and proposed approach shown in this paper will guide the community to develop OOD detection techniques for a practical deployment application.

\bibliographystyle{splncs04}
\bibliography{ref}

\begin{thebibliography}{10}
\providecommand{\url}[1]{\texttt{#1}}
\providecommand{\urlprefix}{URL }
\providecommand{\doi}[1]{https://doi.org/#1}

\bibitem{bendale2016towards}
Bendale, A., Boult, T.E.: Towards open set deep networks. In: Proceedings of
  the IEEE conference on computer vision and pattern recognition. pp.
  1563--1572 (2016)

\bibitem{budhwant2020open}
Budhwant, P., Shinde, S., Ingalhalikar, M.: Open-set recognition for skin
  lesions using dermoscopic images. In: International Workshop on Machine
  Learning in Medical Imaging. pp. 614--623. Springer (2020)

\bibitem{chawla2002smote}
Chawla, N.V., Bowyer, K.W., Hall, L.O., Kegelmeyer, W.P.: Smote: synthetic
  minority over-sampling technique. Journal of artificial intelligence research
   \textbf{16},  321--357 (2002)

\bibitem{chen2021adversarial}
Chen, G., Peng, P., Wang, X., Tian, Y.: Adversarial reciprocal points learning
  for open set recognition. arXiv preprint arXiv:2103.00953  (2021)

\bibitem{codella2018skin}
Codella, N.C., Gutman, D., Celebi, M.E., Helba, B., Marchetti, M.A., Dusza,
  S.W., Kalloo, A., Liopyris, K., Mishra, N., Kittler, H., et~al.: Skin lesion
  analysis toward melanoma detection: A challenge at the 2017 international
  symposium on biomedical imaging (isbi), hosted by the international skin
  imaging collaboration (isic). In: 2018 IEEE 15th international symposium on
  biomedical imaging (ISBI 2018). pp. 168--172. IEEE (2018)

\bibitem{combalia2020uncertainty}
Combalia, M., Hueto, F., Puig, S., Malvehy, J., Vilaplana, V.: Uncertainty
  estimation in deep neural networks for dermoscopic image classification. In:
  Proceedings of the IEEE/CVF Conference on Computer Vision and Pattern
  Recognition Workshops. pp. 744--745 (2020)

\bibitem{esteva2017dermatologist}
Esteva, A., Kuprel, B., Novoa, R.A., Ko, J., Swetter, S.M., Blau, H.M., Thrun,
  S.: Dermatologist-level classification of skin cancer with deep neural
  networks. nature  \textbf{542}(7639),  115--118 (2017)

\bibitem{gal2016dropout}
Gal, Y., Ghahramani, Z.: Dropout as a bayesian approximation: Representing
  model uncertainty in deep learning. In: international conference on machine
  learning. pp. 1050--1059. PMLR (2016)

\bibitem{ge2017generative}
Ge, Z., Demyanov, S., Chen, Z., Garnavi, R.: Generative openmax for multi-class
  open set classification. arXiv preprint arXiv:1707.07418  (2017)

\bibitem{geng2020recent}
Geng, C., Huang, S.j., Chen, S.: Recent advances in open set recognition: A
  survey. IEEE transactions on pattern analysis and machine intelligence
  \textbf{43}(10),  3614--3631 (2020)

\bibitem{gessert2020skin}
Gessert, N., Nielsen, M., Shaikh, M., Werner, R., Schlaefer, A.: Skin lesion
  classification using ensembles of multi-resolution efficientnets with meta
  data. MethodsX  \textbf{7},  100864 (2020)

\bibitem{he2016deep}
He, K., Zhang, X., Ren, S., Sun, J.: Deep residual learning for image
  recognition. In: Proceedings of the IEEE conference on computer vision and
  pattern recognition. pp. 770--778 (2016)

\bibitem{hendrycks2018deep}
Hendrycks, D., Mazeika, M., Dietterich, T.: Deep anomaly detection with outlier
  exposure. arXiv preprint arXiv:1812.04606  (2018)

\bibitem{hendrycks2019using}
Hendrycks, D., Mazeika, M., Kadavath, S., Song, D.: Using self-supervised
  learning can improve model robustness and uncertainty. Advances in Neural
  Information Processing Systems  \textbf{32} (2019)

\bibitem{kim2021out}
Kim, H., Tadesse, G.A., Cintas, C., Speakman, S., Varshney, K.:
  Out-of-distribution detection in dermatology using input perturbation and
  subset scanning. arXiv preprint arXiv:2105.11160  (2021)

\bibitem{krizhevsky2009learning}
Krizhevsky, A., Hinton, G., et~al.: Learning multiple layers of features from
  tiny images  (2009)

\bibitem{lee2017training}
Lee, K., Lee, H., Lee, K., Shin, J.: Training confidence-calibrated classifiers
  for detecting out-of-distribution samples. arXiv preprint arXiv:1711.09325
  (2017)

\bibitem{liang2017enhancing}
Liang, S., Li, Y., Srikant, R.: Enhancing the reliability of
  out-of-distribution image detection in neural networks. arXiv preprint
  arXiv:1706.02690  (2017)

\bibitem{liu2020energy}
Liu, W., Wang, X., Owens, J., Li, Y.: Energy-based out-of-distribution
  detection. Advances in Neural Information Processing Systems  \textbf{33},
  21464--21475 (2020)

\bibitem{liu2019large}
Liu, Z., Miao, Z., Zhan, X., Wang, J., Gong, B., Yu, S.X.: Large-scale
  long-tailed recognition in an open world. In: Proceedings of the IEEE/CVF
  Conference on Computer Vision and Pattern Recognition. pp. 2537--2546 (2019)

\bibitem{mahbod2020transfer}
Mahbod, A., Schaefer, G., Wang, C., Dorffner, G., Ecker, R., Ellinger, I.:
  Transfer learning using a multi-scale and multi-network ensemble for skin
  lesion classification. Computer Methods and Programs in Biomedicine
  \textbf{193},  105475 (2020)

\bibitem{pacheco2020out}
Pacheco, A.G., Sastry, C.S., Trappenberg, T., Oore, S., Krohling, R.A.: On
  out-of-distribution detection algorithms with deep neural skin cancer
  classifiers. In: Proceedings of the IEEE/CVF Conference on Computer Vision
  and Pattern Recognition Workshops. pp. 732--733 (2020)

\bibitem{roy2022does}
Roy, A.G., Ren, J., Azizi, S., Loh, A., Natarajan, V., Mustafa, B., Pawlowski,
  N., Freyberg, J., Liu, Y., Beaver, Z., et~al.: Does your dermatology
  classifier know what it doesn’t know? detecting the long-tail of unseen
  conditions. Medical Image Analysis  \textbf{75},  102274 (2022)

\bibitem{6365193}
Scheirer, W.J., de~Rezende~Rocha, A., Sapkota, A., Boult, T.E.: Toward open set
  recognition. IEEE Transactions on Pattern Analysis and Machine Intelligence
  \textbf{35}(7),  1757--1772 (2013). \doi{10.1109/TPAMI.2012.256}

\bibitem{sun2020conditional}
Sun, X., Yang, Z., Zhang, C., Ling, K.V., Peng, G.: Conditional gaussian
  distribution learning for open set recognition. In: Proceedings of the
  IEEE/CVF Conference on Computer Vision and Pattern Recognition. pp.
  13480--13489 (2020)

\bibitem{tack2020csi}
Tack, J., Mo, S., Jeong, J., Shin, J.: Csi: Novelty detection via contrastive
  learning on distributionally shifted instances. Advances in neural
  information processing systems  \textbf{33},  11839--11852 (2020)

\bibitem{thulasidasan2019mixup}
Thulasidasan, S., Chennupati, G., Bilmes, J.A., Bhattacharya, T., Michalak, S.:
  On mixup training: Improved calibration and predictive uncertainty for deep
  neural networks. Advances in Neural Information Processing Systems
  \textbf{32} (2019)

\bibitem{tschandl2018ham10000}
Tschandl, P., Rosendahl, C., Kittler, H.: The ham10000 dataset, a large
  collection of multi-source dermatoscopic images of common pigmented skin
  lesions. Scientific data  \textbf{5}(1), ~1--9 (2018)

\bibitem{vaze2021open}
Vaze, S., Han, K., Vedaldi, A., Zisserman, A.: Open-set recognition: A good
  closed-set classifier is all you need. arXiv preprint arXiv:2110.06207
  (2021)

\bibitem{yang2018robust}
Yang, H.M., Zhang, X.Y., Yin, F., Liu, C.L.: Robust classification with
  convolutional prototype learning. In: Proceedings of the IEEE conference on
  computer vision and pattern recognition. pp. 3474--3482 (2018)

\bibitem{yoshihashi2019classification}
Yoshihashi, R., Shao, W., Kawakami, R., You, S., Iida, M., Naemura, T.:
  Classification-reconstruction learning for open-set recognition. In:
  Proceedings of the IEEE/CVF Conference on Computer Vision and Pattern
  Recognition. pp. 4016--4025 (2019)

\bibitem{zhang2017mixup}
Zhang, H., Cisse, M., Dauphin, Y.N., Lopez-Paz, D.: mixup: Beyond empirical
  risk minimization. arXiv preprint arXiv:1710.09412  (2017)

\bibitem{zhang2020does}
Zhang, L., Deng, Z., Kawaguchi, K., Ghorbani, A., Zou, J.: How does mixup help
  with robustness and generalization? In: International Conference on Learning
  Representations (2020)

\end{thebibliography}

\end{document}